\documentclass[11pt]{article}
\bibliography{references}
\usepackage{amsmath, amssymb}
\usepackage{graphicx}
\usepackage{tikz}
\usetikzlibrary{arrows.meta, positioning, calc}
\usepackage{geometry}
\usepackage[colorlinks=true, linkcolor=blue, citecolor=blue, urlcolor=blue]{hyperref}
\usepackage[font=small,labelfont=bf]{caption}
\title{From Understanding the World to Intervening in It: A Unified Multi-Scale Framework for Embodied Cognition}
\author{Maijunxian Wang\\
University of California, Davis\\
\texttt{mjxwang@ucdavis.com}}
\date{}

\begin{document}
\maketitle

\begin{abstract}
In this paper, we propose AUKAI, an Adaptive Unified Knowledge-Action Intelligence for embodied cognition that seamlessly integrates perception, memory, and decision-making via multi-scale error feedback. Interpreting AUKAI as an embedded world model \cite{LeCun2022}, our approach simultaneously predicts state transitions and evaluates intervention utility. The framework is underpinned by rigorous theoretical analysis drawn from convergence theory \cite{RobbinsMonro1951}, optimal control \cite{Bellman1957,Kirk2004}, and Bayesian inference \cite{Gelman2013}, which collectively establish conditions for convergence, stability, and near-optimal performance. Furthermore, we present a hybrid implementation that combines the strengths of neural networks with symbolic reasoning modules, thereby enhancing interpretability and robustness \cite{Garcez2015,Besold2017}. Finally, we demonstrate the potential of AUKAI through a detailed application in robotic navigation and obstacle avoidance, and we outline comprehensive experimental plans to validate its effectiveness in both simulated and real-world environments.
\end{abstract}


\newpage
\section{Introduction}
Achieving Artificial General Intelligence (AGI) requires integrating perception, memory, and decision-making into a unified framework. Traditional approaches often treat these components separately, hindering the development of robust, adaptive systems \cite{SuttonBarto2018}. Inspired by seminal works such as Yann LeCun's world model approach \cite{LeCun2022} and Marcus Hutter's AIXI framework \cite{Hutter2005}, our paper introduces the AUKAI framework---a unified model that embeds a world model into an agent to jointly predict state transitions (``knowing'') and evaluate intervention utility (``acting''). Our framework leverages multi-scale error feedback and integrates both neural and symbolic methods to ensure convergence, stability, and near-optimal performance.

Our contributions are:
\begin{itemize}
    \item A novel unified framework (AUKAI) for AGI that integrates perception, memory, and decision-making through multi-scale error feedback.
    \item Rigorous mathematical analysis demonstrating convergence, asymptotic optimality, and stability based on reinforcement learning \cite{SuttonBarto2018}, optimal control \cite{Bellman1957,Kirk2004}, and Bayesian inference \cite{Gelman2013}.
    \item Integration of symbolic reasoning with neural network components for explicit causal inference and uncertainty handling \cite{Garcez2015,Besold2017}.
    \item A comprehensive discussion on parameter relationships and space definitions, as well as operational modes and architecture design (including dual-stream vs.\ single-stream) with practical guidance.
    \item An illustrative multi-scale example in robotic navigation and obstacle avoidance.
\end{itemize}

The remainder of the paper is organized as follows. Section~\ref{sec:related} reviews related work and explains its purpose. Section~\ref{sec:unified} introduces parameter relationships, space definitions, and the unified framework methodology. Section~\ref{sec:arch} discusses architecture design and operational modes. Section~\ref{sec:convergence} provides a detailed convergence and stability analysis. Section~\ref{sec:hybrid} presents the hybrid implementation integrating neural and symbolic components. Section~\ref{sec:multiscale} presents a multi-scale example. Finally, Sections~\ref{sec:experiments} and \ref{sec:discussion} cover experiments, and discussion and conclusion, respectively.

\newpage
\section{Related Work}\label{sec:related}
In the quest for Artificial General Intelligence (AGI), many researchers have pursued methods to integrate perception, memory, and decision-making. Traditional approaches tend to address these components separately, leading to fragmented systems that may excel at individual tasks but fail to generalize robustly across diverse environments \cite{SuttonBarto2018}. In this section, we review the relevant literature in several key areas that inform our work, and we highlight the gaps that the AUKAI framework aims to fill.

\subsection{Perception and Representation Learning}
Deep learning has revolutionized perception tasks, with convolutional neural networks (CNNs) and, more recently, transformers achieving state-of-the-art performance in image and video processing \cite{Krizhevsky2012,Dosovitskiy2020}. In the context of AGI, the challenge is not only to extract robust features from high-dimensional inputs but also to generate low-dimensional representations that capture essential information for further reasoning. Autoencoders and variational autoencoders (VAEs) have been widely used for this purpose \cite{Kingma2013}. However, most methods focus on single-scale representations, whereas many real-world tasks require capturing information at multiple scales.

\subsection{Memory and Temporal Dynamics}
For modeling temporal dependencies, recurrent neural networks (RNNs) and their variants, such as Long Short-Term Memory networks (LSTMs) and Gated Recurrent Units (GRUs), have been extensively studied \cite{Hochreiter1997,Cho2014}. More recently, transformers have also been applied to sequential data due to their capacity to capture long-range dependencies \cite{Vaswani2017}. While these methods excel at modeling time-series data, they often do so at a single temporal scale, which may be insufficient for tasks that require both immediate and long-term planning. Hybrid architectures combining memory networks with hierarchical structures have been proposed, yet a unified theoretical framework that integrates multi-scale temporal memory with decision-making remains lacking.

\subsection{Reinforcement Learning and Decision-Making}
Reinforcement learning (RL) provides a principled approach for decision-making by learning policies through trial-and-error interactions with an environment \cite{SuttonBarto2018}. Methods ranging from Q-learning \cite{Watkins1989} to more recent actor-critic models and Monte Carlo tree search-based algorithms like MuZero \cite{Schrittwieser2020} have achieved impressive results in games and control tasks. However, most RL approaches focus on optimizing a reward signal without explicitly integrating rich perceptual and memory representations. Moreover, many RL algorithms require vast amounts of data and may struggle with stability, particularly in high-dimensional or non-stationary environments.

\subsection{Multi-Scale Modeling and Error Feedback}
In complex environments, information exists at multiple scales. Multi-scale modeling has been successfully applied in computer vision and robotics to capture both local details and global context \cite{Lin2017,Chen2017}. In the realm of embodied cognition, multi-scale error feedback can help reconcile discrepancies between predicted and observed outcomes at various levels of abstraction. Although multi-scale architectures have been proposed in certain domains, they are often implemented as ad-hoc extensions without a rigorous, unified theoretical foundation that connects perception, memory, and decision-making.

\subsection{Neural-Symbolic Integration}
There is a growing interest in combining neural networks with symbolic reasoning to leverage the strengths of both paradigms. Neural methods provide powerful pattern recognition capabilities, whereas symbolic approaches excel in explicit logical reasoning and causal inference \cite{Garcez2015,Besold2017}. Recent work has explored various forms of neural-symbolic integration, such as integrating Bayesian networks or causal graphs with deep learning models. Despite these advances, a comprehensive framework that seamlessly integrates neural and symbolic components into a multi-scale, closed-loop system remains an open challenge.

\subsection{Theoretical Foundations: Convergence, Stability, and Optimal Control}
A rigorous theoretical analysis is essential for ensuring the stability and reliability of complex AI systems. The Bellman equation and contraction mapping principles provide the foundation for convergence analysis in RL \cite{Bellman1957}. Optimal control theory, including the Hamilton-Jacobi-Bellman (HJB) equation, offers further guarantees on optimality in continuous settings \cite{Kirk2004}. Bayesian inference, with its capacity to update beliefs in the presence of uncertainty, has been widely adopted to enhance model robustness \cite{Gelman2013}. However, these theoretical frameworks are rarely applied in an integrated manner that addresses the interplay between multi-scale modeling, memory dynamics, and decision-making.

\subsection{Summary and Gap Analysis}
While significant progress has been made in perception, memory modeling, decision-making, multi-scale processing, and neural-symbolic integration, existing approaches remain fragmented. Most studies focus on isolated components without providing a unified framework that simultaneously addresses:
\begin{itemize}
    \item The integration of multi-scale representations in both spatial and temporal domains.
    \item The incorporation of error feedback mechanisms to guide joint optimization across perception, memory, and decision modules.
    \item The seamless fusion of neural and symbolic methods to enhance interpretability and causal reasoning.
    \item A rigorous theoretical analysis that guarantees convergence and stability of the overall system.
\end{itemize}
The AUKAI framework aims to fill these gaps by providing a unified model that embeds a world model within the agent, leverages multi-scale error feedback, and integrates both neural and symbolic components. In doing so, it establishes a rigorous theoretical foundation based on reinforcement learning, optimal control, and Bayesian inference.

\newpage
\section{Framework Architecture and Methodology}\label{sec:unified}
In this section, we define the parameter spaces and relationships critical for optimizing our unified objective function. We then detail the AUKAI framework, including the unified model formulation, symbol definitions, loss function design, gradient update mechanism, and overall system operation \cite{SuttonBarto2018}.

\subsection{Parameter Spaces and Relationships}
\subsubsection*{Space Definitions}
\begin{itemize}
    \item \(\mathcal{X}\): Observation space (e.g., raw images, sensor readings) \cite{SuttonBarto2018}.
    \item \(\mathcal{S}\): State space. \(s_t \in \mathcal{S}\) is a lower-dimensional, continuous representation extracted from the preprocessed input \(\tilde{x}_t\) \cite{Bellman1957}.
    \item \(\mathcal{H}\): Memory space, where \(h_t \in \mathcal{H}\) encodes temporal dependencies \cite{Hochreiter1997}.
    \item \(\mathcal{A}\): Action space, which can be discrete or continuous \cite{Watkins1989}.
    \item \(\Theta\): Overall parameter space, defined as:
    \[
    \Theta = \Phi \times \Theta_{\text{mem}} \times \Theta_{\text{wm}},
    \]
    where:
    \begin{itemize}
        \item \(\Phi\) is the parameter space of the perception network \cite{Kingma2013}.
        \item \(\Theta_{\text{mem}}\) is the parameter space of the memory module \cite{Hochreiter1997}.
        \item \(\Theta_{\text{wm}}\) is the parameter space of the world model and causal reasoning module \cite{Gelman2013}.
    \end{itemize}
\end{itemize}

Let \(\theta = (\phi, \theta_{\text{mem}}, \theta_{\text{wm}})\) denote the overall parameters. The optimal parameter set \(\theta^*\) is obtained by minimizing the overall loss function \cite{RobbinsMonro1951}.

\subsection{Unified Model and Overall Objective}
We construct an embedded world model:
\[
P_\theta(s_{t+1}, u_t \mid s_t, a_t),
\]
where:
\begin{itemize}
    \item \(P_\theta(s_{t+1}\mid s_t, a_t)\) predicts the next state distribution \cite{Bellman1957}.
    \item \(u_t = U(s_t,a_t;\theta)\) evaluates the intervention utility.
\end{itemize}
This embedded framework draws inspiration from reinforcement learning principles and optimal control theory, aiming to balance accurate prediction of environmental dynamics with effective action selection.\cite{LeCun2022} The objective is to find the optimal parameter set that minimizes the expected discounted sum of prediction errors while maximizing intervention utility \cite{SuttonBarto2018}. Formally, the objective is defined as:
\[
\theta^* = \arg\min_{\theta} \; \mathbb{E}\left[\sum_{t=0}^{\infty}\gamma^t \Bigl( L_{\text{pred}}(s_t,a_t,s_{t+1};\theta) - \beta\, U(s_t,a_t;\theta) \Bigr)\right],
\]
with discount factor \(\gamma\) and utility weight \(\beta\) \cite{SuttonBarto2018,RobbinsMonro1951}.

\subsection{Symbol, Parameter, and Loss Definitions}
\subsubsection*{Perception}
\begin{itemize}
    \item \(x_t\): Raw observation.
    \item \(\tilde{x}_t\): Preprocessed input.
    \item \(s_t = g_\phi(\tilde{x}_t)\): State representation.
    \item \(L_{\text{perception}} = \| \tilde{x}_t - g_\phi^{-1}(s_t) \|^2\): Reconstruction loss, as commonly used in autoencoder frameworks \cite{Kingma2013}.
\end{itemize}

\subsubsection{Memory}
\begin{itemize}
    \item \(h_t = f_{\text{memory}}(s_t, h_{t-1}; \theta_{\text{mem}})\): Memory module output \cite{Hochreiter1997}.
    \item \(h_t^{target}\): Target state from \(f_{\text{memory}}^{\text{target}}(s_t, h_{t-1})\).
    \item \(L_{\text{memory}} = \| h_t - h_t^{target} \|^2\): Memory consistency loss.
\end{itemize}

\subsubsection{World Model and Intervention Utility}
\begin{itemize}
    \item \(\hat{s}_{t+1} = P_\theta(s_{t+1}\mid h_t,a_t)\): Predicted next state.
    \item \(L_{\text{pred}}(s_t,a_t,s_{t+1};\theta)=D_{KL}\Bigl(P_{\text{real}}(s_{t+1}\mid s_t,a_t) \,\Big\|\, P_\theta(s_{t+1}\mid s_t,a_t)\Bigr)\): Prediction loss, where the Kullback-Leibler divergence is employed as a measure of distribution discrepancy \cite{Gelman2013}.
    \item \(U(s_t,a_t;\theta)\): Intervention utility.
    \item \(\beta\): Utility weight.
\end{itemize}

\subsubsection{Multi-Scale Extension}
For \(l \in \{\text{micro}, \text{meso}, \text{macro}\}\) with weights \(w_l\) (\(\sum_l w_l=1\)):
\[
P_\theta(s_{t+1}\mid s_t,a_t)=\sum_l w_l\,P_{\theta,l}(s_{t+1}\mid s_t,a_t),
\]
\[
L_{\text{pred}}(s_t,a_t,s_{t+1};\theta)=\sum_l w_l\,D_{KL}\Bigl(P_{\text{real},l}(s_{t+1}\mid s_t,a_t) \,\Big\|\, P_{\theta,l}(s_{t+1}\mid s_t,a_t)\Bigr).
\]
This multi-scale formulation is inspired by recent advances in hierarchical modeling \cite{Chen2017,Lin2017}.

\subsubsection{Decision Making}
\[
a_t^* = \arg\max_{a\in\mathcal{A}} U(s_t,a;\theta),
\]
and in multi-scale:
\[
U(s_t,a_t;\theta)=\sum_l w_l\,U_l(s_t^l,a_t;\theta).
\]
The action selection mechanism follows standard reinforcement learning practices \cite{Watkins1989}.

\subsection{Overall Loss Function and Gradient Updates}
The overall loss is defined as:
\[
\mathcal{L} = \sum_{t=0}^{\infty} \gamma^t \Bigl( L_{\text{perception}} + L_{\text{memory}} + L_{\text{pred}} - \beta\, U(s_t,a_t;\theta) + L_{\text{aux}} \Bigr),
\]
where \(L_{\text{aux}}\) represents auxiliary losses (e.g., regularization) \cite{SuttonBarto2018}.

The gradient updates for each module are:
\begin{itemize}
    \item \textbf{Perception Network:}
    \[
    \frac{\partial \mathcal{L}}{\partial \phi} = \frac{\partial L_{\text{perception}}}{\partial s_t}\frac{\partial s_t}{\partial \phi} + \frac{\partial L_{\text{pred}}}{\partial s_t}\frac{\partial s_t}{\partial \phi}.
    \]
    \item \textbf{Memory Module:}
    \[
    \frac{\partial \mathcal{L}}{\partial \theta_{\text{mem}}} = \frac{\partial L_{\text{pred}}}{\partial h_t}\frac{\partial h_t}{\partial \theta_{\text{mem}}} + \frac{\partial L_{\text{memory}}}{\partial h_t}\frac{\partial h_t}{\partial \theta_{\text{mem}}}.
    \]
    \item \textbf{World Model \& Causal Reasoning Module:}
    \[
    \frac{\partial \mathcal{L}}{\partial \theta_{\text{wm}}} = \frac{\partial L_{\text{pred}}}{\partial \hat{s}_{t+1}}\frac{\partial \hat{s}_{t+1}}{\partial \theta_{\text{wm}}}.
    \]
\end{itemize}
The overall parameter update is:
\[
\theta \leftarrow \theta - \eta\, \nabla_\theta \Bigl(L_{\text{pred}}(s_t,a_t,s_{t+1};\theta) - \beta\, U(s_t,a_t;\theta)\Bigr),
\]
which is derived following standard stochastic gradient descent principles \cite{RobbinsMonro1951}.

\subsection{System Operation and Closed-Loop Update}
At each time step \(t\), the system performs:
\begin{enumerate}
    \item \textbf{Perception:} Preprocess \(x_t\) to \(\tilde{x}_t\) and compute \(s_t = g_\phi(\tilde{x}_t)\).
    \item \textbf{Memory Update:} Update \(h_t = f_{\text{memory}}(s_t, h_{t-1}; \theta_{\text{mem}})\).
    \item \textbf{Prediction and Utility Evaluation:} Compute \(\hat{s}_{t+1} = P_\theta(s_{t+1}\mid h_t,a_t)\) and \(U(s_t,a_t;\theta)\).
    \item \textbf{Decision and Interaction:} Choose \(a_t^* = \arg\max_{a\in\mathcal{A}} U(s_t,a;\theta)\) and execute it; the environment returns the true \(s_{t+1}\) (extracted from \(x_{t+1}\)) \cite{Watkins1989}.
    \item \textbf{Parameter Update:} Update parameters using:
    \[
    \theta \leftarrow \theta - \eta\, \nabla_\theta \Bigl(L_{\text{pred}}(s_t,a_t,s_{t+1};\theta) - \beta\, U(s_t,a_t;\theta)\Bigr),
    \]
    following gradient descent rules \cite{RobbinsMonro1951}.
\end{enumerate}

\textbf{Figure~\ref{fig:aici_improved} below illustrates the closed-loop operation, showing both forward data flow and key error feedback paths} \cite{Chen2017,Lin2017}.

\begin{figure}[p]
    \centering
    \caption{AUKAI Framework: Forward flow (solid) and key error feedback (dashed).}
    \label{fig:aici_improved}
    \begin{tikzpicture}[
      node distance=1.8cm, 
      auto, 
      >=Latex,
      every node/.style={draw, rectangle, align=center, rounded corners},
      ]
        % 1. External World (top): W_t
        \node (world1) {External World\\ \(W_t\)\\ \(x_t \in \mathcal{X}\)};
        
        % 2. Preprocessing
        \node (preproc) [below=of world1] {Preprocessing\\ (Denoising, Normalization)};
        
        % 3. Perceptual Module (S)
        \node (perception) [below=of preproc] {Perceptual Module (S)\\ \(s_t = g_\phi(\tilde{x}_t)\)\\ \(s_t \in \mathcal{S}\)\\ \(L_{\text{perception}}\)};
        
        % 4. Memory Module (M)
        \node (memory) [below=of perception] {Memory Module (M)\\ \(h_t = f_{\text{memory}}(s_t, h_{t-1}; \theta_{\text{mem}})\in \mathcal{H}\)\\ Target: \(h_t^{target} \in \mathcal{H}\)\\ \(L_{\text{memory}}\)};
        
        % 5. World Model \& Causal Reasoning (WM+R)
        \node (wm) [below=2.2cm of memory] {World Model \& Causal Reasoning (WM+R)\\ \(\hat{s}_{t+1} = P_\theta(s_{t+1}\mid h_t,a)\)\\ \(U(h_t,a; \theta)\), \(L_{\text{pred}}\)};
        
        % 6. Action Module (A)
        \node (action) [below=of wm] {Action Module (A)\\ \(a_t^* = \arg\max_{a\in\mathcal{A}} U(h_t,a; \theta)\)};
        
        % 7. Next External World (W_{t+1})
        \node (world2) [below=of action] {External World\\ \(W_{t+1}\)\\ Real Observation \(x_{t+1}\)};
    
        % -- Main forward flow (solid lines) --
        \draw[->] (world1) -- (preproc) node[midway, right] {\(x_t\)};
        \draw[->] (preproc) -- (perception) node[midway, right] {\(\tilde{x}_t\)};
        \draw[->] (perception) -- ( memory) node[midway, right] {\(s_t\)};
        \draw[->] (memory.south) -- ++(0,-1.0) -| (wm.north) node[midway, right] {\(h_t\)};
        \draw[->] (wm) -- (action) node[midway, right] {Pred. \(\hat{s}_{t+1}\) \\ \& Utility \(U\)};
        \draw[->] (action) -- (world2) node[midway, right] {\(a_t^*\)};
    
        % -- Dashed feedback arrows --
        \draw[->, dashed]
          (world2.east) -- ++(3,3)
          node[right, align=center]{Real Perception\\ \(s_{t+1}\in \mathcal{S}\)}
        |- ($(wm.east)+(0,-0.2)$);
    
        \draw[->, dashed] 
          (wm.west) -- ++(0,4.1)
          node[left, align=center]{WM Pred. Error +\\ \(L_{\text{memory}} + L_{\text{aux}}\)\\ \(\Rightarrow\) update \(\theta_{\text{mem}}\)}
          |- ($(memory.west)+(0,-0.2)$);
    
        \draw[->, dashed] 
          (wm.west) -- ++(0,7.9)
          node[left, align=center]{WM Pred. Error +\\ \(L_{\text{perception}} + L_{\text{aux}}\)\\ \(\Rightarrow\) update \(\phi\)}
          |- ($(perception.west)+(0,-0.2)$);
    \end{tikzpicture}
\end{figure}
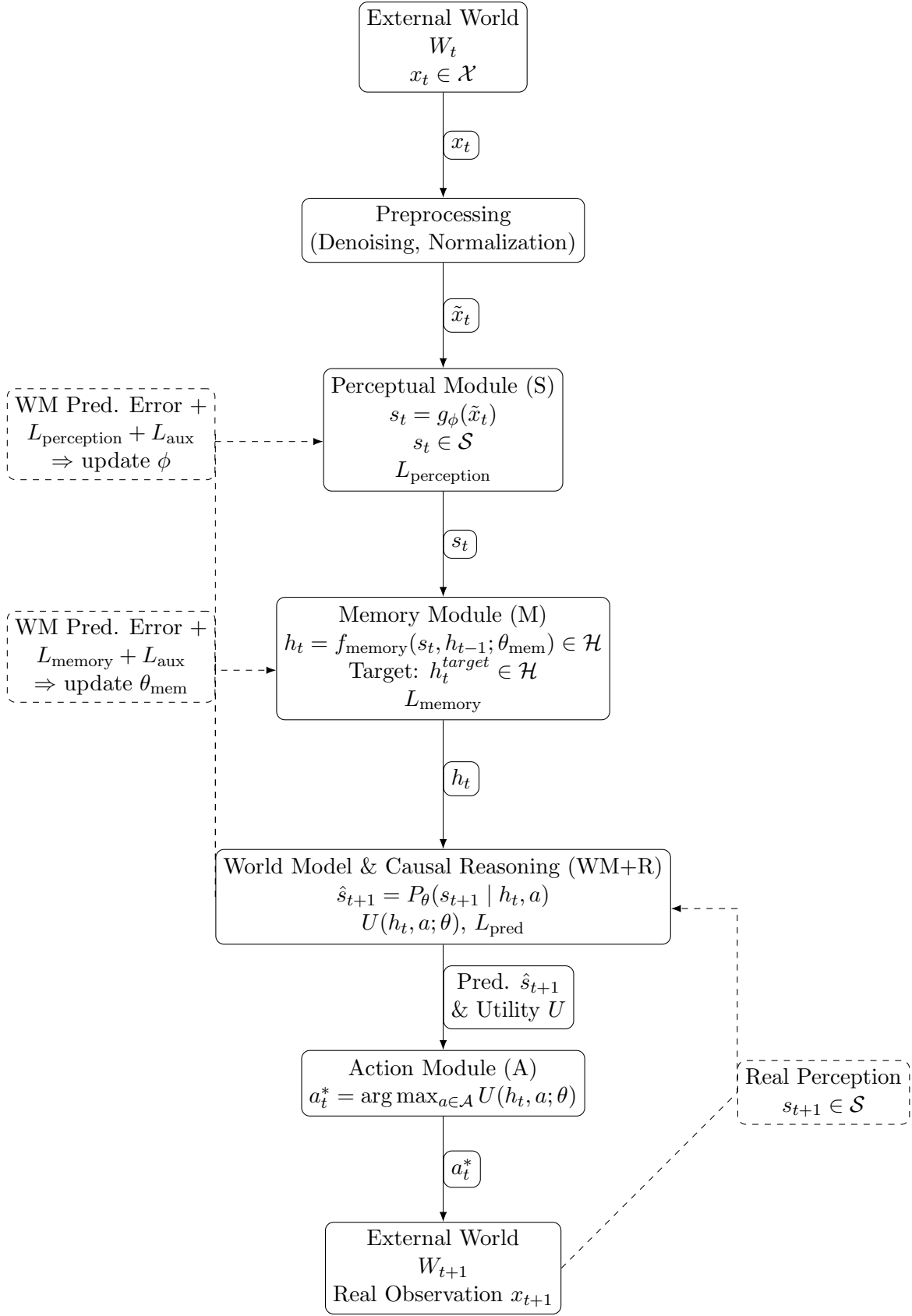

\newpage
\section{Architecture Design and Operational Modes}\label{sec:arch}
This section discusses various operational modes and architectural choices, which have been explored in related work in reinforcement learning and hybrid system design \cite{SuttonBarto2018,Schrittwieser2020}.

\subsection{Operational Modes}
AUKAI can operate in different modes:
\begin{itemize}
    \item \textbf{Single-Flow (Modeling-Only Mode):} The system focuses on environmental modeling:
    \[
    x_t \rightarrow S \rightarrow M \rightarrow WM+R \rightarrow \text{Feedback: Compare } \hat{s}_{t+1} \text{ with } s_{t+1}.
    \]
    \item \textbf{Single-Flow (Intervention-Only Mode):} Uses a pre-trained model for decision-making:
    \[
    x_t \rightarrow S \rightarrow M \ (\text{minimal update}) \rightarrow WM+R \rightarrow A \rightarrow \text{Execute } a_t^*.
    \]
    \item \textbf{Dual-Flow (Integrated Mode):} Both modeling and intervention flows operate concurrently:
    \[
    \begin{array}{c}
    \text{Modeling Flow: } x_t \rightarrow S \rightarrow M \rightarrow WM+R \rightarrow \hat{s}_{t+1} \rightarrow \text{Update}, \\
    \text{Intervention Flow: } WM+R \rightarrow A \rightarrow a_t^* \rightarrow \text{Feedback } x_{t+1}.
    \end{array}
    \]
\end{itemize}

\subsection{Dual-Stream vs. Single-Stream Architectures}
\begin{itemize}
    \item \textbf{Dual-Stream Architecture:} Separates into:
    \begin{itemize}
        \item \textbf{Knowledge Stream:} \(s_t^{(k)} = g_\phi^{(k)}(\tilde{x}_t)\)
        \item \textbf{Action Stream:} \(s_t^{(a)} = g_\phi^{(a)}(\tilde{x}_t)\)
    \end{itemize}
    They are fused with memory output:
    \[
    z_t = \text{Fusion}(s_t^{(k)}, s_t^{(a)}, h_t).
    \]
    \item \textbf{Single-Stream Architecture:} Processes all information jointly:
    \[
    z_t = f_{\text{single}}(\tilde{x}_t, h_{t-1};\theta).
    \]
\end{itemize}

\subsection{Discussion}
The AUKAI framework supports multiple operational modes to cater to diverse application requirements. In the Single-Flow (Modeling-Only Mode), the system prioritizes accurate environmental modeling by processing inputs through perception and memory modules, then predicting future states and comparing them with actual observations. In contrast, the Single-Flow (Intervention-Only Mode) leverages a pre-trained model to rapidly select and execute actions, with minimal updates to the model, making it suitable for real-time decision-making. The Dual-Flow (Integrated Mode) simultaneously maintains a continuous environmental model while performing action selection, balancing robust model refinement with immediate responsiveness. These operational strategies are grounded in established practices in reinforcement learning and hybrid planning \cite{SuttonBarto2018,Schrittwieser2020}, providing practical guidance for system design tailored to specific application needs.

\newpage
\section{Convergence and Stability Analysis}\label{sec:convergence}
In this section, we present a more detailed and rigorous analysis of the convergence and stability properties of the AUKAI framework. Our analysis leverages concepts from dynamic programming, optimal control, and stochastic approximation theory \cite{Bellman1957,Kirk2004,RobbinsMonro1951}.

\subsection{Convergence Analysis via Contraction Mapping}
We consider the prediction module within a Markov Decision Process (MDP) framework. Let \(V(s)\) denote a value function over states, and let the Bellman operator \(T\) be defined as
\[
(TV)(s) = \min_{a \in \mathcal{A}} \left\{ L_{\text{pred}}(s, a, s') + \gamma V(s') \right\},
\]
where \(s'\) is drawn from \(P_\theta(s'|s,a)\) and \(\gamma \in (0,1)\) is the discount factor.

\paragraph{Contraction Property.}  
Under standard conditions, the Bellman operator \(T\) is a contraction mapping in the sup-norm:
\[
\|TV_1 - TV_2\|_\infty \leq \gamma \|V_1 - V_2\|_\infty,
\]
for any two value functions \(V_1\) and \(V_2\). By the Banach fixed-point theorem, there exists a unique fixed point \(V^*\) such that
\[
V^* = T V^*,
\]
and iteratively applying \(T\) (i.e., performing value iteration) converges to \(V^*\) \cite{Bellman1957}.

\paragraph{Implication for AUKAI.}  
In our framework, the prediction loss \(L_{\text{pred}}\) is designed to mimic the Bellman error. With a properly chosen learning rate schedule (satisfying \cite{RobbinsMonro1951}), the gradient descent updates on \(\theta\) are analogous to performing a contraction mapping iteration on the induced value function. Thus, the overall optimization process converges to a fixed point \(\theta^*\), provided that the following conditions hold:
\begin{enumerate}
    \item The learning rate \(\eta_t\) satisfies \(\sum_t \eta_t = \infty\) and \(\sum_t \eta_t^2 < \infty\) \cite{RobbinsMonro1951}.
    \item There is sufficient exploration so that all state-action pairs are visited infinitely often \cite{SuttonBarto2018}.
\end{enumerate}

\subsection{Stability Analysis via Lyapunov Methods}
To analyze the stability of the closed-loop system, we construct a Lyapunov function \(V(x)\), where \(x\) denotes the overall system state (including perceptual, memory, and predicted state components).

\paragraph{Lyapunov Condition.}  
We require that for the system update
\[
x_{t+1} = f(x_t, a_t, \theta),
\]
there exists a positive definite function \(V(x)\) and a positive definite function \(W(x)\) such that
\[
V(x_{t+1}) - V(x_t) \leq -W(x_t) \quad \forall x_t.
\]
This inequality guarantees that the system state converges asymptotically to an equilibrium point (i.e., the fixed point corresponding to \(\theta^*\)) \cite{Kirk2004}.

\paragraph{Application to AUKAI.}  
In our framework, the error feedback (including both the prediction loss and auxiliary losses) drives the system state toward a desired equilibrium. If each component loss is convex (or locally convex) and the overall loss \(\mathcal{L}\) serves as a Lyapunov function candidate, then the updates ensure that:
\[
\mathcal{L}(x_{t+1}) - \mathcal{L}(x_t) \leq -W(x_t),
\]
which implies asymptotic stability of the closed-loop system.

\subsection{Stochastic Approximation and Robbins-Monro Conditions}
Since the optimization is performed in a stochastic setting (due to the sampling of observations and state transitions), we interpret our parameter updates as an instance of stochastic approximation. Under the Robbins-Monro conditions, if:
\begin{itemize}
    \item The noise in the gradient estimates is bounded.
    \item The learning rate schedule \(\{\eta_t\}\) satisfies \(\sum_{t=1}^\infty \eta_t = \infty\) and \(\sum_{t=1}^\infty \eta_t^2 < \infty\) \cite{RobbinsMonro1951},
\end{itemize}
then the stochastic gradient descent (SGD) process converges almost surely to a stationary point of the expected loss function.

\subsection{Handling Gradient Conflicts}
In multi-task and modular systems, conflicting gradients can impede convergence. To mitigate this issue, methods such as Projected Conflicting Gradient (PCGrad) are employed to resolve gradient interference by projecting conflicting gradients onto a common descent direction. This ensures that the updates from different loss components (e.g., \(L_{\text{pred}}\), \(L_{\text{memory}}\), \(L_{\text{perception}}\), and \(L_{\text{aux}}\)) do not counteract each other, thus preserving the overall contractive behavior and ensuring stable convergence \cite{Yu2020}.

\subsection{Summary}
Combining the contraction mapping argument, Lyapunov stability analysis, and stochastic approximation theory, we conclude that under standard assumptions, the AUKAI framework converges to a unique fixed point \(\theta^*\) and maintains stability during training. These rigorous theoretical foundations validate the effectiveness of our design choices in achieving robust and adaptive embodied cognition.

\newpage
\section{Hybrid Implementation: Integrating Neural and Symbolic Components}\label{sec:hybrid}
To enhance interpretability and robustness, AUKAI can be implemented as a hybrid system:
\begin{enumerate}
    \item \textbf{Neural Components:} Deep networks (e.g., CNNs, RNNs, Transformers) extract features and model temporal dependencies \cite{Vaswani2017}.
    \item \textbf{Symbolic Components:} Symbolic reasoning modules (e.g., Bayesian networks, causal graphs) perform explicit causal inference and update a knowledge base \cite{Garcez2015,Besold2017}.
    \item \textbf{Fusion and Decision:} The neural output \(s_t\) and memory \(h_t\) are supplemented by corrective feedback \(\Delta s_t\) from the symbolic module. The fused representation is:
    \[
    z_t = \text{Fusion}(s_t, h_t, \Delta s_t),
    \]
    and the optimal action is chosen as:
    \[
    a_t^*=\arg\max_{a\in\mathcal{A}} U(z_t,a;\theta).
    \]
    \item \textbf{Feedback Update:} After executing \(a_t^*\) and receiving \(s_{t+1}\), the neural components are updated via backpropagation, while the symbolic module is updated using Bayesian inference:
    \[
    P(\theta|D)=\frac{P(D|\theta)P(\theta)}{P(D)}.
    \]
    This Bayesian update mechanism is crucial for dynamically adapting the knowledge base as new data emerges \cite{Gelman2013}.
\end{enumerate}

\subsubsection*{Discussion}
This hybrid approach combines neural methods for high-dimensional data processing with symbolic methods for explicit causal reasoning, thereby enhancing interpretability and uncertainty management. Symbolic modules—such as Bayesian networks or causal graphs—excel at inferring cause-effect relationships, enabling the framework to better understand the environment's underlying structure and making decision-making more robust under uncertainty. Neural models, on the other hand, are highly effective at feature extraction and handling noisy data, while symbolic methods offer clear logical reasoning and enforce domain constraints. Recent studies also discuss the challenges of neural-symbolic integration \cite{Yu2021} and emphasize the need for novel methods to overcome interoperability issues. Furthermore, the symbolic component can be updated via Bayesian inference to dynamically adapt its knowledge base as new data emerges \cite{Gelman2013}. However, coordinating the gradient updates from the neural side with the probabilistic updates of the symbolic side is challenging and misalignment may lead to suboptimal updates or slow convergence. Additionally, the integration of symbolic reasoning may introduce extra computational overhead and scalability issues that require careful optimization.

\newpage
\section{Multi-Scale Example: Robotic Navigation and Obstacle Avoidance}\label{sec:multiscale}
We present a multi-scale example in robotic navigation and obstacle avoidance.

\subsection{Micro Scale (Low Scale)}
\begin{itemize}
    \item \textbf{State Extraction:} \(s_t^{\text{micro}} = g_\phi^{\text{micro}}(\tilde{x}_t^{\text{micro}})\).
    \item \textbf{State Prediction:} \(\hat{s}_{t+1}^{\text{micro}} = P_{\theta,\text{micro}}(s_{t+1}^{\text{micro}} \mid s_t^{\text{micro}}, a_t)\).
    \item \textbf{Utility:} \(U_{\text{micro}}(s_t^{\text{micro}}, a_t;\theta)\).
    \item \textbf{Loss:} \(L_{\text{pred, micro}} = \frac{1}{2}\| s_{t+1}^{\text{micro}} - \hat{s}_{t+1}^{\text{micro}} \|^2\).
\end{itemize}

\subsection{Meso Scale (Intermediate Scale)}
\begin{itemize}
    \item \textbf{State Extraction:} \(s_t^{\text{meso}} = g_\phi^{\text{meso}}(\{ \tilde{x}_{t-\tau}^{\text{micro}}, \dots, \tilde{x}_t^{\text{micro}}\})\).
    \item \textbf{State Prediction:} \(\hat{s}_{t+1}^{\text{meso}} = P_{\theta,\text{meso}}(s_{t+1}^{\text{meso}} \mid s_t^{\text{meso}}, a_t)\).
    \item \textbf{Utility:} \(U_{\text{meso}}(s_t^{\text{meso}}, a_t;\theta)\).
    \item \textbf{Loss:}
    \[
    L_{\text{pred, meso}} = D_{KL}\Bigl(P_{\text{real, meso}}(s_{t+1}^{\text{meso}} \mid s_t^{\text{meso}}, a_t) \,\Big\|\, P_{\theta,\text{meso}}(s_{t+1}^{\text{meso}} \mid s_t^{\text{meso}}, a_t)\Bigr).
    \]
\end{itemize}

\subsection{Macro Scale (High Scale)}
\begin{itemize}
    \item \textbf{State Extraction:} \(s_t^{\text{macro}} = g_\phi^{\text{macro}}(M_t)\), where \(M_t\) is global information (e.g., a map).
    \item \textbf{State Prediction:} \(\hat{s}_{t+1}^{\text{macro}} = P_{\theta,\text{macro}}(s_{t+1}^{\text{macro}} \mid s_t^{\text{macro}}, a_t)\).
    \item \textbf{Utility:} \(U_{\text{macro}}(s_t^{\text{macro}}, a_t;\theta)\).
    \item \textbf{Loss:} \(L_{\text{pred, macro}} = \frac{1}{2}\| s_{t+1}^{\text{macro}} - \hat{s}_{t+1}^{\text{macro}} \|^2\).
\end{itemize}

The overall multi-scale loss and utility are:
\[
L_{\text{pred}} = w_{\text{micro}}\, L_{\text{pred, micro}} + w_{\text{meso}}\, L_{\text{pred, meso}} + w_{\text{macro}}\, L_{\text{pred, macro}},
\]
\[
U(s_t,a_t;\theta) = w_{\text{micro}}\, U_{\text{micro}}(s_t^{\text{micro}}, a_t;\theta) + w_{\text{meso}}\, U_{\text{meso}}(s_t^{\text{meso}}, a_t;\theta) + w_{\text{macro}}\, U_{\text{macro}}(s_t^{\text{macro}}, a_t;\theta).
\]
Parameters are updated via:
\[
\theta \leftarrow \theta - \eta\, \nabla_\theta \Bigl( L_{\text{pred}}(s_t,a_t,s_{t+1};\theta) - \beta\, U(s_t,a_t;\theta) \Bigr).
\]

\newpage
\section{Experiments}\label{sec:experiments}
\textbf{Note on Experiments:} It is important to emphasize that the primary focus of this paper is the formulation and theoretical analysis of the AUKAI framework. The experiments described in this section serve as a preliminary outline of our planned empirical evaluations and are intended to illustrate potential avenues for validating the framework in both simulated and real-world environments. In subsequent research, we will conduct detailed experiments to further verify the performance, robustness, and practical applicability of AUKAI. For now, these experimental descriptions should be understood as an outline of future work rather than final, conclusive results.

To validate the feasibility and effectiveness of the AUKAI framework, we propose a series of comprehensive experiments designed to assess its performance in both simulated and real-world environments. Our experimental plan is divided into several components:

\subsection{Simulation Experiments}
In the simulation phase, we will evaluate the framework on standard control and navigation tasks:
\begin{itemize}
    \item \textbf{Benchmark Control Tasks:} We will test AUKAI on classical reinforcement learning benchmarks such as CartPole, MountainCar, and Acrobot. These tasks will help us assess the convergence properties and stability of the multi-scale prediction and decision-making modules \cite{SuttonBarto2018}.
    \item \textbf{Simulated Navigation:} We will create simulated environments with varying complexity (e.g., maze-like environments, dynamic obstacle scenarios) to test the framework’s ability to perform robotic navigation and obstacle avoidance. This will allow us to evaluate how well the multi-scale error feedback mechanism reconciles information across different spatial and temporal resolutions.
    \item \textbf{Ablation Studies in Simulation:} We will systematically remove or modify key components (e.g., multi-scale feedback, memory module, symbolic integration) to analyze their individual contributions to the overall system performance.
\end{itemize}

\subsection{Real-World Data Experiments}
To demonstrate the framework’s applicability in realistic settings, we plan to conduct experiments using publicly available datasets:
\begin{itemize}
    \item \textbf{Autonomous Driving Datasets:} We will leverage datasets such as KITTI \cite{Geiger2012} and Waymo \cite{Sun2020} to validate the performance of AUKAI in complex, real-world navigation tasks. This will include testing the framework's perception and decision-making modules under diverse environmental conditions.
    \item \textbf{Robotic Manipulation:} Additionally, we plan to utilize datasets from robotic manipulation tasks to assess how AUKAI integrates perception and memory for precise control in cluttered environments.
\end{itemize}

\subsection{Ablation Studies}
We will conduct detailed ablation studies to isolate and evaluate the contribution of each component within the AUKAI framework:
\begin{itemize}
    \item \textbf{Multi-Scale vs. Single-Scale:} Compare the performance of the full multi-scale architecture against a baseline single-scale version to quantify the benefits of incorporating multiple spatial and temporal scales.
    \item \textbf{Hybrid vs. Pure Neural Implementation:} Assess the impact of the symbolic reasoning module by comparing the hybrid AUKAI framework with a version that relies solely on neural network components.
    \item \textbf{Gradient Conflict Resolution:} Evaluate the effectiveness of techniques such as PCGrad ~\cite{Yu2020} in mitigating gradient conflicts within our multi-task learning setting.
\end{itemize}

\subsection{Comparative Studies}
To further establish the merits of the AUKAI framework, we will compare its performance with state-of-the-art methods:
\begin{itemize}
    \item \textbf{Baseline Reinforcement Learning Methods:} Compare against standard RL algorithms such as DQN~\cite{Mnih2015}, PPO~\cite{Schulman2017}, and MuZero~\cite{Schrittwieser2020} on metrics such as convergence speed, sample efficiency, and stability.
    \item \textbf{Hybrid and Neural-Symbolic Systems:} Recent studies have highlighted challenges in neural-symbolic integration and stressed the need for novel methods to overcome interoperability issues~\cite{Yu2021}.Benchmark AUKAI against recent hybrid frameworks that combine neural and symbolic methods to evaluate improvements in interpretability, decision robustness, and uncertainty management.
\end{itemize}

\subsection{Evaluation Metrics}
Across all experiments, we will use a variety of quantitative and qualitative metrics. Standard evaluation metrics for convergence, stability, and uncertainty management have been widely adopted in reinforcement learning literature~\cite{Hessel2018}:
\begin{itemize}
    \item \textbf{Convergence and Stability Metrics:} Monitor the evolution of the loss functions, the variance of gradient updates, and the speed of convergence.
    \item \textbf{Control Performance:} Measure performance in control tasks via standard metrics such as cumulative reward, task completion rate, and time-to-goal.
    \item \textbf{Robustness and Adaptability:} Evaluate the system’s performance under perturbations, noise, and non-stationary conditions.
    \item \textbf{Interpretability and Uncertainty:} Assess the quality of symbolic reasoning outputs and the framework’s ability to manage uncertainty in decision-making.
\end{itemize}

Through this comprehensive set of experiments, we aim to demonstrate that the AUKAI framework not only advances the theoretical understanding of embodied cognition but also offers practical advantages in terms of performance, robustness, and interpretability in both simulated and real-world applications.

\newpage
\section{Discussion and Conclusion}\label{sec:discussion}
We discuss the advantages, limitations, and future directions:

In this section, we combine our discussion and conclusion to provide an integrated overview of our work.

\textbf{Theoretical Guarantees:}\\
Our analyses based on reinforcement learning, optimal control, and Bayesian inference have established rigorous guarantees for convergence, asymptotic optimality, and stability~\cite{Bellman1957, RobbinsMonro1951, SuttonBarto2018, Gelman2013}. These results reinforce the credibility of the AUKAI framework and its potential to function reliably in dynamic and uncertain environments. However, these guarantees are derived under specific assumptions, which future work should aim to relax for broader applicability.

\textbf{Unified Multi-Scale and Modular Design:}\\
A significant strength of our framework is its unified design that integrates multi-scale models, intervention utility, and error feedback into a single optimization objective. This holistic integration aligns the processes of perception, memory, and decision-making, thereby addressing the limitations inherent in modular approaches. Nevertheless, balancing the contributions from different scales remains challenging and warrants further investigation. In future work, exploring adaptive weighting strategies for multi-scale components \cite{Deshpande2016} could further enhance performance.

\textbf{Hybrid Implementation:}\\
By combining neural networks with symbolic reasoning, our framework leverages the strengths of both paradigms. Neural networks excel in pattern recognition and learning, while symbolic methods enhance causal inference and interpretability~\cite{Garcez2015, Besold2017}. Recent studies have also discussed the challenges of neural-symbolic integration, including issues such as gradient conflicts and interoperability \cite{Yu2021}. Addressing these challenges through improved integration techniques is an important direction for future research.

\textbf{Parameter Spaces and Relationships:}\\
Clearly defining the observation, state, memory, and action spaces, along with the overall parameter space, provides a comprehensive view of the optimization process. This detailed specification facilitates parameter tuning and contributes to the system’s robustness and interpretability. However, the high dimensionality and complexity of these spaces might require further dimensionality reduction or advanced optimization strategies, particularly in large-scale environments.

\textbf{Architecture and Operational Modes:}\\
Our framework supports multiple operational modes (modeling-only, intervention-only, and dual-flow integrated modes) and includes a comparative analysis of dual-stream versus single-stream architectures. This discussion offers practical insights for system design tailored to specific application needs. Future studies should explore alternative configurations to further enhance performance under diverse conditions.

\textbf{Future Directions and Conclusion:}\\
Looking ahead, several research avenues are promising. We plan to investigate adaptive weighting strategies for sub-models to dynamically balance their contributions \cite{Luo2021, Perez2023}, develop advanced gradient conflict resolution techniques to harmonize learning between neural and symbolic components, and conduct extensive experimental validations using real-world datasets and diverse control tasks. Continuous refinement of the neural-symbolic integration will further improve both interpretability and overall system performance. In summary, while the AUKAI framework represents a promising advance toward robust embodied cognition in artificial agents, addressing the aforementioned challenges is crucial for its further development and practical deployment.

\end{document}